\documentclass{article}
\usepackage{amsmath,graphicx}
\usepackage{authblk}

\usepackage{float}
\usepackage{multirow}
\usepackage{mathrsfs,bm}
\newtheorem{property}{Property}
\usepackage{amsfonts,dsfont,eucal}
\usepackage[capitalize]{cleveref}
\usepackage{color}
\usepackage{subfigure}
\usepackage{siunitx}
\usepackage{array,booktabs}

\DeclareMathOperator*{\argmin}{arg\,min}

\providecommand{\mat}[1]{{\bm{#1}}} %
\providecommand{\ve}[1]{{\bm{#1}}} %

\newcommand{\Real}{\mathbb{R}}

\newcommand\redsout{\bgroup\markoverwith{\textcolor{red}{\rule[0.5ex]{2pt}{0.4pt}}}\ULon}

\title{The Representation Jensen-Rényi Divergence}
\author[1]{Jhoan Keider {Hoyos Osorio}}
\author[1]{Oscar Skean}
\author[2]{Austin J. Brockmeier}
\author[1]{Luis Gonzalo {Sanchez Giraldo}}
\affil[1]{University of Kentucky}
\affil[2]{University of Delaware}
\date{}                     
\begin{document}
%
\maketitle
\begin{abstract}
We introduce a divergence measure between data distributions based on operators in reproducing kernel Hilbert spaces defined by kernels.  The empirical estimator of the divergence is computed using the eigenvalues of positive definite Gram matrices that are obtained by evaluating the kernel over pairs of data points. The new measure shares similar properties to Jensen-Shannon divergence.  Convergence of the proposed estimators follows from concentration results based on the difference between the ordered spectrum of the Gram matrices and the integral operators associated with the population quantities. The proposed measure of divergence avoids the estimation of the probability distribution underlying the data. Numerical experiments involving comparing distributions and applications to sampling unbalanced data for classification show that the proposed divergence can achieve state of the art results.
\end{abstract}
\begin{keywords}
Divergence, Kernel methods, Information theoretic learning, Subsampling, Imbalanced classification
\end{keywords}
\section{Introduction}
\label{sec:intro}

Estimating divergences from empirical data is a problem that finds several applications in machine learning and signal processing. Following previous work \cite{sanchez2015ieeetransit}, we propose an alternative non-parametric measure of divergence with desirable convergence properties that avoids estimating the distribution of the data.   
Current approaches based on neural networks frame the problem of divergence estimation as a variational problem~\cite{nowozin2016fgan}, where the estimator of divergence is obtained by training a neural network to estimate a lower bound. Other approaches~\cite{seth2011,gretton2012kernel,zhang2019optimal}, including the one proposed here, rely on embedding the distribution in a feature space and treating the estimator of divergence as a statistic in this feature space. A notable example in this line of work is the maximum mean discrepancy (MMD)~\cite{gretton2012kernel}, which represents a distribution as a mean element in a reproducing kernel Hilbert space (RKHS). In particular, when a characteristic kernel is employed \cite{fukumizu2009kernel}, MMD is a distance metric between probability distributions. Here, we propose a divergence measure that is obtained by measuring the dependence between the random variable representing the data point and a random variable labeling the data point's distribution. In this case, the divergence is minimal when conditioning on the label does not change the sampling distribution. To measure this dependence, we employ a recently proposed information theoretic quantity that behaves similar to mutual information and has convergence guarantees close to MMD. Unlike MMD, the proposed measure makes use of the entire spectrum of an uncentered covariance operator in an RKHS, which has the potential to make the measure more sensitive to differences in distributions. The resulting divergence bear some resemblance with quantum divergences (see for instance \cite{sra2021LinearAlgebraApp}). However, the interpretations and the roles of the positive definite matrices are different.
The paper is organized as follows. We start by briefly revisiting the relation between Jensen-Shannon divergence and Shannon's mutual information to motivate the idea of using a measure of mutual information to construct a divergence. Then, we introduce the representation entropy, an information theoretic quantity that has similar properties to R\'{e}nyi's entropy to construct a measure of mutual information from which we obtain our proposed measure of divergence. We then discuss a few properties of the proposed divergence as well as some statistical guarantees. Finally, we test the proposed divergence and compare results with state-of-the-art methods.

\section{Background}
Informally, we can say that two distributions differ if we can can determine the originating distribution of a point randomly drawn from either. Consider two random variables $X$ and $L$, where $X\in \mathcal{X}$ is a continuous random variable and $L$ is Bernoulli. Let $p(x|L=0)$  and $p(x|L=1)$ be the two conditional densities. $p(x|L=0) =  p(x|L=1), \forall x \in\mathcal{X}$ is equivalent to saying that $X$ and $L$ are statistically independent, and, furthermore, $ P(L = l|x) = P(L = l)$, for $l \in \{0,1\}$. In other words, observing $X=x$ provides no information about the conditional distribution. This idea has been exploited in the context of generative adversarial networks (GANs)~\cite{goodfellow2014gan}, where the accuracy of the discriminator is a measure of how different the real and generated data distributions are. We can make this intuition clearer by revisiting a well-known result in information theory that relates the Jensen-Shannon (JS) divergence with Shannon's mutual information.\\
The JS divergence is a symmetrized version of the Kullback-Liebler divergence. Its square root is also a distance metric between probability distributions \cite{endres2003ieeetransit}. Let $(\mathcal{X}, \mathcal{B}_{\mathcal{X}})$ be a measurable space and $P$ and $Q$ be two probability measures defined on it, with $M=\frac{1}{2}P+\frac{1}{2}Q$. The JS divergence is
\begin{equation}\label{eq:js_divergence}
{\rm JS}(P \Vert Q) = \frac{1}{2}\int_{\mathcal{X}} \log{\left(\frac{dP}{dM}\right)}dP + \frac{1}{2} \int_{\mathcal{X}} \log{\left(\frac{dQ}{dM}\right)}dQ
\end{equation}
Alternatively, the JS divergence is the mutual information between a random variable $Z$ sampled from the mixture of the two distributions, $Z \sim M$, with a Bernoulli random variable $L \sim \mathrm{Ber}(\frac{1}{2})$ indicating which distribution.  $Z|_{L = 0} \sim P$, $Z|_{L = 1} \sim Q$. It is straightforward to show that the mutual information $I(Z;L) = H(Z) - H(Z|L) =  H(Z) - \frac{1}{2}[H(Z|L\!=\!0) + H(Z|L\!=\!1)] = {\rm JS}(P \parallel Q)$.
Here, we propose a divergence based on a novel measure of mutual information that 
we call ``representation mutual information''. This measure has similar properties to Shannon's mutual information, but is based on the more general R\'{e}nyi entropy instead of Shannon entropy, and can be estimated from data without making strong assumptions about the data distribution. 
\subsection{Representation Entropy and Mutual Information}\label{sec:representation_entropy_mi}
The representation entropy, also known as matrix-based entropy, is analogous to R\'{e}nyi's $\alpha$-order entropy. Let $\kappa : \mathcal{X} \times \mathcal{X} \rightarrow \mathbb{R}_{\geq 0}$ be a positive definite kernel. For a sample $\left\{\mathbf{x}_i \right\}_{i=1}^N \subset \mathcal{X}$ of $N$ objects drawn from an unknown probability distribution $P$, let $\mathbf{K}$ be an $N \times N$ normalized Gram matrix with entries $\left(\mathbf{K}\right)_{ij} = \frac{\kappa(\mathbf{x}_i, \mathbf{x}_j)}{\sqrt{\kappa(\mathbf{x}_i, \mathbf{x}_i),\kappa(\mathbf{x}_j, \mathbf{x}_j)}}$. The $\alpha$-order representation entropy is defined as
\begin{equation}\label{eq:representation_entropy}
S_{\alpha}(\mathbf{K}) = \frac{1}{1 - \alpha}\log{\left(\mathrm{tr}\left[\left(\frac{1}{N}\mathbf{K}\right)^{\alpha}\right]\right)}.
\end{equation}

Unlike plug-in estimators that approximate the probability distribution as an intermediate step, representation entropy can be understood as a statistic in a reproducing kernel Hilbert space. Based on the Hadamard product, representation entropy can be extended to joint entropy of two random variables $X \in \mathcal{X}$ and $Y \in \mathcal{Y}$ as follows. Let $\kappa_{X}$ and $\kappa_{Y}$ be kernels defined on $\mathcal{X}\times\mathcal{X}$ and $\mathcal{Y}\times\mathcal{Y}$, respectively. For a joint sample $\left\{(\mathbf{x}_i, \mathbf{y}_i)\right\}_{i=1}^N$, let $\left(\mathbf{K}_X\right)_{ij} = \frac{\kappa_X(\mathbf{x}_i, \mathbf{x}_j)}{\sqrt{\kappa_X(\mathbf{x}_i, \mathbf{x}_i),\kappa_X(\mathbf{x}_j, \mathbf{x}_j)}}$ and $\left(\mathbf{K}_Y\right)_{ij} = \frac{\kappa_Y(\mathbf{x}_i, \mathbf{x}_j)}{\sqrt{\kappa_Y(\mathbf{x}_i, \mathbf{x}_i),\kappa_Y(\mathbf{x}_j, \mathbf{x}_j)}}$. The representation joint entropy of order $\alpha$ is defined as:
\begin{equation}\label{eq:representation_joint_entropy}
S_{\alpha}(\mathbf{K}_X, \mathbf{K}_Y) = \frac{1}{1 - \alpha}\log{\left[\mathrm{tr}\left[\left(\frac{1}{N}\mathbf{K}_X \circ \mathbf{K}_Y\right)^{\alpha}\right]\right]}.
\end{equation}
Additionally, it can be shown that $S_{\alpha}(\mathbf{K}_X, \mathbf{K}_Y) \geq S_{\alpha}(\mathbf{K}_X)$ and $S_{\alpha}(\mathbf{K}_X, \mathbf{K}_Y)$ $\leq S_{\alpha}(\mathbf{K}_X) + S_{\alpha}(\mathbf{K}_Y)$. Based on the second inequality, we define 
the $\alpha$-order representation mutual information as
\begin{equation}\label{eq:representation_mutual_information}
I_{\alpha}(\mathbf{K}_Y ; \mathbf{K}_X) = S_{\alpha}(\mathbf{K}_X) + S_{\alpha}(\mathbf{K}_Y) - S_{\alpha}(\mathbf{K}_X, \mathbf{K}_Y).
\end{equation}

\section{The Representation Jensen-R\'{e}nyi Divergence}
Here we consider measuring the divergence directly between two samples $\left\{\mathbf{x}_i\right\}_{i=1}^N \subset \mathcal{X}$ and  $\left\{\mathbf{y}_i\right\}_{i=1}^M \subset \mathcal{X}$ in the same space $\mathcal{X}$. Let $\mathbf{X} = \left\{\mathbf{x}_i\right\}_{i=1}^N$ and  $\mathbf{Y} = \left\{\mathbf{y}_i\right\}_{i=1}^M$. The mixture sample $\mathbf{Z} = \left\{\mathbf{z}_i\right\}_ {i=1}^{N+M}$ is defined as $\mathbf{z}_i = \mathbf{x}_i$ for $i\in \{1,\dots,N\}$ and $\mathbf{z}_i = \mathbf{y}_{i-N}$ for $i\in \{N+1, \dots, N+M\}$. The variable indicating the origin of each point is $\mathbf{l}_i = 0$ for $i\in\{1,\dots,N\}$ and $\mathbf{l}_i = 1$ for $i\in \{N+1, \dots, N+M\}$.
Given the kernel $\kappa$, and a normalized version $\tilde{\kappa}$, the mixture Gram matrix is $
\left(\mathbf{K}_Z\right)_{ij}=\tilde{\kappa}(\mathbf{z}_i,\mathbf{z}_j)=\frac{\kappa(\mathbf{z}_i,\mathbf{z}_j)}{\sqrt{\kappa(\mathbf{z}_i,\mathbf{z}_i)\kappa(\mathbf{z}_j,\mathbf{z}_j)} }  
$, and the entries of the indicator Gram matrix $\mathbf{L}$ are obtained using the 0-1 kernel as $
\left(\mathbf{L}\right)_{ij} = \left\{\begin{array}{lc}
0 & \textrm{if}\: \mathbf{l}_i \neq \mathbf{l}_j \\
1 & \textrm{if}\: \mathbf{l}_i = \mathbf{l}_j   
\end{array} \right.$.
We define the representation Jensen-R\'{e}nyi divergence (JRD), $\mathrm{D}_{\alpha}(\mathbf{X} \Vert \mathbf{Y})$ in terms of \eqref{eq:representation_mutual_information} as
\begin{equation}\label{eq:representation_jr_divergence}
\mathrm{D}_{\alpha}(\mathbf{X} \Vert \mathbf{Y}) = I_{\alpha}(\mathbf{K}_Z ; \mathbf{L}).
\end{equation}
\subsection{Properties of the Representation JRD}
\begin{property} $D_{\alpha}$ is symmetric. 
\begin{equation}
\mathrm{D}_{\alpha}(\mathbf{X} \Vert  \mathbf{Y}) = \mathrm{D}_{\alpha}(\mathbf{Y} \Vert  \mathbf{X}).
\end{equation}
\end{property}
\begin{property}
For any $\mathbf{X}_{\Pi} = \left\{\mathbf{x}_{\Pi_i}\right\}_{i=1}^N$ obtained by a permutation $\Pi$ of the elements in sample set $\mathbf{X} = \left\{\mathbf{x}_i\right\}_{i=1}^N$, 
\begin{equation}
\mathrm{D}_{\alpha}(\mathbf{X} \Vert \mathbf{X}_{\Pi}) = 0.
\end{equation}
\end{property}
\begin{property}  $D_{\alpha}$ is upper bounded by
\begin{equation}
\mathrm{D}_{\alpha}(\mathbf{X}\Vert \mathbf{Y}) \leq \frac{1}{1 - \alpha}\log{\left[\left(\frac{N}{N+M}\right)^{\alpha}+\left(\frac{M}{N+M}\right)^{\alpha}\right]}.
\end{equation} 
\end{property}

\subsection{Statistical Behavior}
Following \cite{sanchez2015ieeetransit}, we introduce the population quantity associated with \eqref{eq:representation_jr_divergence}. Let $(\mathcal{X}, \mathcal{B}_{\mathcal{X}})$ be a countably generated measure space. Let $P$ and $Q$ be the probability distributions on this space, where $\mathbf{X}$ and $\mathbf{Y}$ are i.i.d. samples from $P$ and $Q$, respectively. The labeled points $\{(\mathbf{z}_i,\mathbf{l}_i)\}_{i=1}^{N+M}$ are i.i.d. samples from the joint distribution $P_{ZL}$. 

For a normalized kernel $\tilde\kappa$, we define the operator $G: \mathcal{H} \mapsto \mathcal{H}$ via the bilinear form:
\begin{equation}\label{eq:cov_operator}
\langle f, G g\rangle = \int\limits_{\mathcal{X}}\langle f, \phi(\mathbf{x})\rangle \langle \phi(\mathbf{x}), g \rangle dM(\mathbf{x}),\quad f,g\in\mathcal{H},
\end{equation}
where $M= \frac{1}{2}P + \frac{1}{2}Q$ and $\phi : \mathcal{X} \rightarrow \mathcal{H}$ is the map to an RKHS induced by $\tilde\kappa$.
For the product kernel $\tilde\kappa(\mathbf{x}, \mathbf{x}^\prime )l(\mathbf{l}, \mathbf{l}^\prime ) = \langle \phi_{\otimes}(\mathbf{x}, \mathbf{l}), \phi_{\otimes}(\mathbf{x}^\prime, \mathbf{l}^\prime)\rangle$, we define the operator $H$ on elements $f_\otimes, g_\otimes \in \mathcal{H}\otimes \mathcal{I}$, where $\mathcal{I}$ is the Hilbert space of the embedded indicator variable, as 
\begin{equation}\label{eq:joint_cov_operator}
\langle f_\otimes, H g_\otimes\rangle = \int\limits_{\mathcal{X}, \mathcal{L}}\langle f_\otimes, \phi_{\otimes}(\mathbf{x}, \mathbf{l})\rangle \langle \phi_{\otimes}(\mathbf{x}, \mathbf{l}), g_\otimes \rangle dP_{ZL}(\mathbf{x}, \mathbf{l}).
\end{equation}
Then, for $\alpha>1$, the population version of \eqref{eq:representation_jr_divergence} is
\begin{equation}\label{eq:representation_jr_divergence_population}
\mathrm{D}_{\alpha}(P_X \Vert P_Y) = \log{2} - \frac{1}{\alpha-1}\log{ \left[\frac{\mathrm{tr}(G^{\alpha})}{\mathrm{tr}(H^{\alpha})} \right]}.
\end{equation}
 As shown before~\cite{sanchez2015ieeetransit}, with probability $1 - \delta$,
\begin{equation}
\left\vert \frac{\mathrm{tr}(G^{\alpha})}{\mathrm{tr}(H^{\alpha})} - \frac{\mathrm{tr}(\hat{G}^{\alpha})}{\mathrm{tr}(\hat{H}^{\alpha})}\right\vert \leq \frac{2\alpha}{\mathrm{tr}(H^{\alpha})}  \sqrt{\frac{\log{\frac{2}{\delta}}}{N}},
\end{equation} 
where
$ \mathrm{tr}(\hat{G}^{\alpha}) = \mathrm{tr}\left(\frac{1}{2N}\mathbf{K}_Z\right)^{\alpha}$ and $\mathrm{tr}(\hat{H}^{\alpha}) = \mathrm{tr}\left(\frac{1}{2N}\mathbf{K}_Z \circ \mathbf{L}\right)^{\alpha}$ use the empirical counterparts of $G$ and $H$, assuming equal sample sizes of $N=M$, 
\begin{equation}\mathrm{D}_{\alpha}(\mathbf{X} \Vert \mathbf{Y}) =  \log{2} - \frac{1}{\alpha-1}\log{ \left[\frac{\mathrm{tr}(\hat{G}^{\alpha})}{\mathrm{tr}(\hat{H}^{\alpha})} \right]}.
\end{equation}

\subsection{Random Fourier Approximation}
The eigendecomposition of a Gram matrix $\mathbf{K}$ has $\mathcal{O}(N^2)$ memory and $\mathcal{O}(N^3)$ time complexities, which become prohibitive for large sample sizes. We use random Fourier features (RFFs) \cite{rahimi2007random} to approximate the kernel evaluations between two data points $\ve{x}$ , $\ve{y}$ with an explicit feature mapping $\ve{\hat{\phi}}$, that is, $\tilde{\kappa}(\ve{x},\ve{y}) = \langle\,\phi(\ve{x}),\phi(\ve{y})\rangle \approx \ve{\hat{\phi}}(\ve{x})^{\top}\ve{\hat{\phi}}(\ve{y})$.



Let $\mathbf{\hat{\Phi}}_X \in \Real^{N\times D}$ be the matrix containing in the $i$th row the randomized feature mapping of $\mathbf{x}_i$. We can approximate the Gram matrix $\mathbf{K}_X$ by $\mathbf{\hat{\Phi}}_X\mathbf{\hat{\Phi}}_X^{\top}$. Since $\mathbf{\hat{\Phi}}_X\mathbf{\hat{\Phi}}_X^{\top}$ and $\mathbf{\hat{\Phi}}_X^{\top}\mathbf{\hat{\Phi}}_X$ share the same  nonzero eigenvalues, for $N \gg D$,  we can obtain significant reductions in computation if we use $\mathbf{\hat{\Phi}}_X^{\top}\mathbf{\hat{\Phi}}_X$ to approximate the eigenvalues of $\mathbf{K}_X$. 
To approximate the representation JRD, we also need to consider the joint entropy $S_\alpha(\mathbf{K}_Z,\mathbf{L}_Z)$. Since  $\mat{K}_Z \circ \mat{L}_Z$ is block diagonal with $(\mathbf{K}_X)_{ij}=\tilde\kappa(\mathbf{x}_i,\mathbf{x}_j)$ on the left upper block, and $(\mathbf{K}_Y)_{ij}=\tilde\kappa(\mathbf{y}_i,\mathbf{y}_j)$ on the right lower block, $\lambda\left(\frac{1}{N}\mathbf{K}_Z \circ \mathbf{L}_Z\right) = \lambda\left(\frac{1}{N}\mathbf{K}_X\right)\cup \lambda\left(\frac{1}{N}\mathbf{K}_Y\right)\approx \lambda\left(\frac{1}{N}\mathbf{\hat{\Phi}}_X^{\top}\mathbf{\hat{\Phi}}_X\right) \cup \lambda\left(\frac{1}{N}\mathbf{\hat{\Phi}}_Y^{\top}\mathbf{\hat{\Phi}}_Y\right)$.

\section{Experiments}
\subsection{Synthetic Data}
\begin{figure}
    \centering
    \subfigure[]{\includegraphics[height=0.37\linewidth]{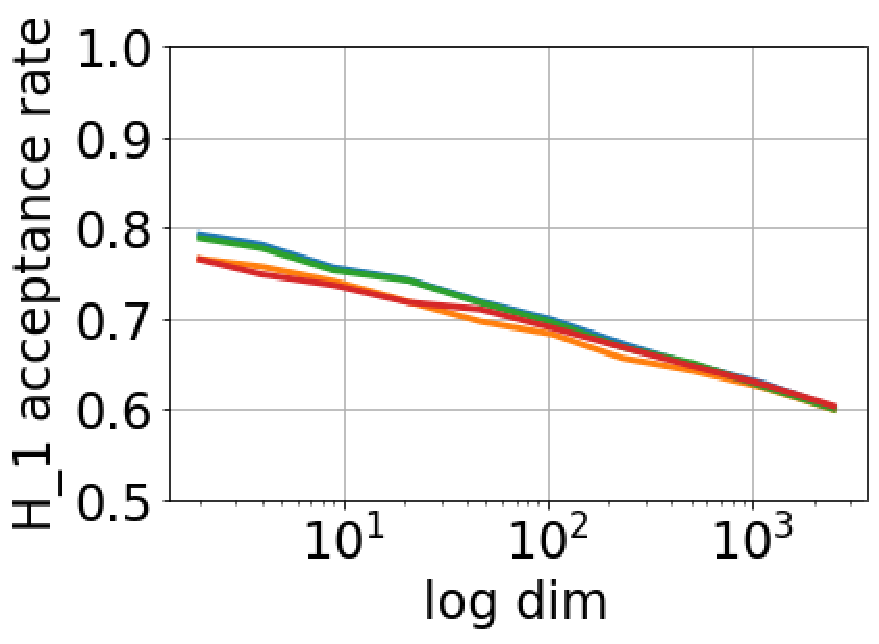}\label{fig:means}}
    \subfigure[]{\includegraphics[height=0.37\linewidth]{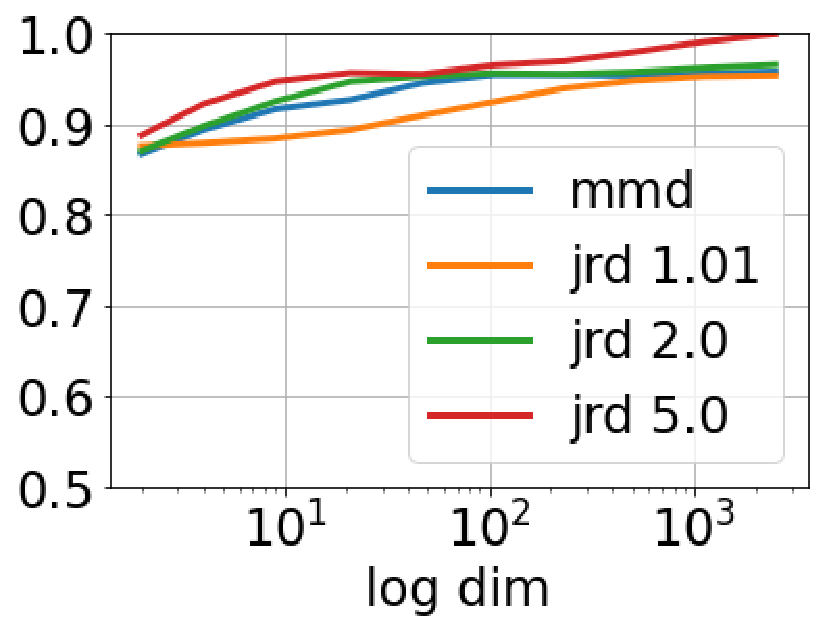}\label{fig:variances}}
    \caption{Statistical power for two-samples tests: MMD and representation JRD. (a) Samples from isotropic Gaussians with equal variance and different means. (b) Isotropic Gaussians with equal means and different variances. }\label{fig:hypothesis_test}
\end{figure}

 To understand how the JRD behaves when detecting differences in distributions, we performed several permutation tests based on the representation divergence between two samples $\mat{X}\in \Real^{N\times d}$ and $\mat{Y} \in \Real^{M\times d}$ with equal sizes $N=M = 250$. In these experiments, we tested the performance of our divergence in terms of the dimensionality $d$ and compared against MMD. The kernel size is set as $\sigma =\sqrt{\frac{1}{2N(2N-1)}\sum_{i <j}||\mathbf{z}_i - \mathbf{z}_j||_2^2}$. 500 random permutations were used to generate a surrogate distribution of the statistic under the null hypothesis. 100 Monte Carlo samples were used to calculate the statistical power.  In the first experiment, we tested two set of samples drawn from Gaussian distributions as proposed in \cite{gretton2012kernel}. Both distributions had unit variance but different means. We tested 20 different Euclidean distances between the means logarithmically spaced from 0.05 to 50, and computed the power for each, averaging the results for a given dimensionality. For the second experiment, both distributions had zero mean and covariance matrices $\bm{\Sigma}_X = \mat{I}$, $\bm{\Sigma}_Y = c \mat{I}$. We tested 20 different values of $c$ logarithmically spaced from $10^{0.01}$ to $10$ and averaged together the results for a given dimensionality. 
We also evaluated the influence of  $\alpha$ in the performance of JRD, namely $\alpha = 1.01$, $\alpha = 2$, and $\alpha = 5$. The significance level for all the permutation tests was $\tau = 0.05$. The results are shown in Figure \ref{fig:hypothesis_test}.  
In this example, we can see the influence of the $\alpha$ parameter. For this example, for $\alpha = 2$ both MMD and JRD work similar. In the case of the difference in variances, $\alpha > 2$ performs better, highlighting the advantage of using the eigenvalues as opposed to the mean in the RKHS.
\subsection{Representation JRD-VAE}
We tested a modified instance of the InfoVAE \cite{ermon2017infovae} that we call JRD-VAE. Similar to the MMD-VAE, the JRD-VAE minimizes a combination of the auto-encoder reconstruction error and the representation JRD between the true prior of the code distribution and the sampled encoding of the data. The enconder network is composed of 2 convolutional layers with 64 and 128 channels followed by 2 fully connected layers with sizes 6272 and 1024. The decoder is a transposed version of the encoding network. For the classification experiments, the dimension of the code space is set to 10. We trained a support vector machine classifier on the codes learned by the JRD-VAE for different values of $\alpha$. We also trained a MMD-VAE baseline model for reference (see \cref{tab:jrd_vae_svm}).
\begin{table}[t]
    \caption{SVM test accuracy for MNIST  trained on the encoding learned by the JRD-VAE with different $\alpha$}
    \label{tab:jrd_vae_svm}
    \centering
    \small
    \begin{tabular}{cccc}
         \textbf{MMD-VAE} & \textbf{JRD 1.01} & \textbf{JRD 2.0} & \textbf{JRD 5.0} \\ \hline
         $0.976$ & $\mathbf{0.9802}$ & $0.9763$ & $0.9778$ \\
    \end{tabular}

\end{table}
In this case, $\alpha =1.01$ performs better that the baseline model. For $\alpha=2$ the algorithms behaves similar to MMD.
\subsection{Imbalanced Data Representation}
We evaluated the JRD as a subsampling technique to balance skewed datasets in the context of imbalanced data classification. In a two-class imbalanced classification dataset, we define the class with the highest number of samples as the majority class, and the other as the minority class. Let  $\mathbf{X}_{+}\in\Real^{N_{+} \times d}$ and $\mathbf{X}_{-}\in\Real^{N_{-} \times d}$ denote the data matrices for the majority and minority classes respectively,  where $N_{+}\gg N_{-}$. We want to subsample $M = N_{-} $ instances from the majority class, to prevent biased classification results. We can use the representation JRD to find a subset $\tilde{\mathbf{X}}_+ \in \Real^{M \times d}$ that minimizes its divergence with the majority class:
\begin{equation}
  \tilde{\mathbf{X}}_+ =  \argmin_{\tilde{\mathbf{X}}^* \in \Real^{M \times d} } \mathrm{D}_{\alpha}(\tilde{\mathbf{X}}^*\Vert\mathbf{X}_{+}).
\end{equation} 
A similar objective based on the Cauchy-Schwartz divergence \cite{principe2010itlbook} to subsample imbalanced datasets was recently introduced \cite{hoyos2021relevant}. However, that method requires tuning of one additional parameter and approaches to scale to large datasets were not investigated. 
In our experiments we use $\alpha = 1.01$ and the Gaussian kernel to compute representation JRD. The kernel bandwidth is set as the median of the Euclidean distance among the majority class samples. We also implemented the JRD subsampling using RFF to approximate the divergence. Specifically, we tested 3 different number of RFFs, $D=256$, $D=512$, and $D=1024$. Finally, to evaluate the subsampling performance in the context of imbalanced data classification, we train a support vector machine (SVM) with Gaussian kernel. To tune the SVM parameters we implement a 5-fold nested cross-validation strategy. The kernel width is searched within the range $[0.01 \sigma_0, 3\sigma_0]$, where $\sigma_0$ is the median of the Euclidean distance between training samples, and the penalty value is tuned within the logspace range $[0,3]$. For comparison purposes we measure the classification performance as the area under the ROC curve (AUC). 

We assessed the JRD subsampling on 44 imbalanced, two-class datasets from the ``Knowledge Extraction based on Evolutionary Learning'' (KEEL) repository \cite{fernandez2008study}. The imbalance ratios range from $1.8$ to $129$. Additionally, we compare the results against six state-of-the-art methods for imbalanced data classification that were tested on this same repository: RUSBoost1 (RUS1) \cite{seiffert2010rusboost,sun2020class}, Underbagging4 (UB4)~\cite{barandela2003new,raghuwanshi2019class}, SMOTEBagging4 (SBAG4)~\cite{wang2009diversity,sun2018imbalanced}, Clustering-based undersampling with AdaBoost (CUS-AB)~\cite{CBUS3}, Relevant information sampling (RIUS), and clustered relevant information sampling (CRIUS) \cite{hoyos2021relevant}. 

\begin{table}[t]
\caption{Comparison of majority subsampling results. Numbers indicate the Jensen-Rényi divergence wins (and defeats) in terms AUC among the state-of-the-art subsampling methods. The last two rows are the results for a sign-test ($\tau = 0.05$) over all datasets.}

\label{tab:imbalanced}
	\resizebox{\linewidth}{!}{
		\renewcommand{\arraystretch}{0.8}
\begin{tabular}{l|cccccc}
\multicolumn{1}{c}{\multirow{2}{*}{}} & \multicolumn{6}{c}{State-of-the-art methods}                                                                                                                             \\ 
\multicolumn{1}{c|}{}                  & \multicolumn{1}{c}{\textbf{RUS1}} & \multicolumn{1}{c}{\textbf{UB4}} & \multicolumn{1}{c}{\textbf{SMOTE}} & \multicolumn{1}{c}{\textbf{CUS-AB}} & \multicolumn{1}{c}{\textbf{RIUS}} & \multicolumn{1}{c}{\textbf{CRIUS}} \\ \hline
\textbf{JRD}                       &  35(0)               &  34(0)              &  33(0)               &  21(1)                      &  14(0)              &  3(0)                \\
\textbf{JRD-RFF (256)}                         &  32(1)               &  28(1)              &  29(1)                &  21(4)                       &  11(2)               &  6(4)                 \\
\textbf{JRD-RFF (512)}                         &  31(1)               &  26(1)              &  29(0)                &  16(3)                       &  15(2)               &  2(5)                 \\
\textbf{JRD-RFF (1024)}                         &  29(1)               &  30(0)              &  29(0)               &  22(3)                       &  11(1)               &  4(4)                \\ \hline
JRD sign test               & 1                        & 1                       & 1                         & 1                                & 1                       & 0                         \\
\textit{p-value}              & 1.5e-8                  &  1.5e-9                 & 1.0e-10                  & 4.7e-3                           & 7.5e-6                 & 9.5e-1 \\   
\end{tabular}
}
\end{table}

 \cref{tab:imbalanced} reports the results of JRD versus the aforementioned methods across the 44 datasets.  JRD outperformed most of the methods, showing superior performance on most of the datasets, although the results were more competitive with the most recent sampling methods. Compared with the regular RIUS, the JRD subsampling seems to have better capability to properly represent  the underlying structure of the majority class since it was statistically superior in 14 out of 44 datasets (1-sample $t$-test with a 5\% significance level). With respect to CRIUS, our method is slightly better, exhibiting superior classification performance on 3 datasets. Nevertheless, note that CRIUS is an ensemble algorithm which employs several classifiers while our method only uses one. Moreover, it is worth mentioning that in the 44 different datasets tested, our method is almost never statistically inferior than any of the 6 tested methods. Even with small number of RFFs, we can get competitive results compared with the state-of-the-art methods. \cref{tab:imbalanced} shows the hypothesis test over all the average performances in the entire repository, where we can see that in general JRD subsampling is statistically better than 5 of the 6 algorithms and competitive with the remaining one. 


\section{Conclusions}
We introduced a measure of divergence based on representation entropy. The proposed measure has desirable convergence properties and can be tuned to be more sensitive to the difference in distributions. Several examples show that  representation JRD can be better than MMD at detecting discrepancies between distributions. However, such advantages come from selecting the proper $\alpha$ parameter, that controls the influence of the eigenvalues involved in the computation of the JRD. Different example applications highlight the potential use of the proposed divergence for different problems in machine learning ranging from representation learning to correcting data imbalances. 

\section{Acknowledgments}

This material is based upon work supported by the Office of the Under Secretary of Defense for Research and Engineering under award number FA9550-21-1-0227. Austin Brockmeier's effort were sponsored by the Department of the Navy, Office of Naval Research under ONR award number N00014-21-1-2300.


\label{sec:refs}

\bibliographystyle{IEEEbib}
\bibliography{refs}

\begin{thebibliography}{10}

\bibitem{sanchez2015ieeetransit}
Luis~Gonzalo Sanchez~Giraldo, Murali Rao, and Jose~C. Principe,
\newblock ``Measures of entropy from data using infinitely divisible kernels,''
\newblock {\em IEEE Transactions on Information Theory}, vol. 61, no. 1, pp.
  535--548, 2015.

\bibitem{nowozin2016fgan}
Sebastian Nowozin, Botond Cseke, and Ryota Tomioka,
\newblock ``f-{GAN}: Training generative neural samplers using variational
  divergence minimization,''
\newblock in {\em Advances in Neural Information Processing Systems 29}, D.~D.
  Lee, M.~Sugiyama, U.~V. Luxburg, I.~Guyon, and R.~Garnett, Eds., pp.
  271--279. Curran Associates, Inc., 2016.

\bibitem{seth2011}
Sohan Seth, Austin~J. Brockmeier, and José~C. Príncipe,
\newblock ``A metric approach toward point process divergence,''
\newblock in {\em 2011 IEEE International Conference on Acoustics, Speech and
  Signal Processing (ICASSP)}, 2011, pp. 2104--2107.

\bibitem{gretton2012kernel}
Arthur Gretton, Karsten~M. Borgwardt, Malte~J. Rasch, Bernhard Sch{\"o}lkopf,
  and Alexander Smola,
\newblock ``A kernel two-sample test,''
\newblock {\em The Journal of Machine Learning Research}, vol. 13, no. 1, pp.
  723--773, 2012.

\bibitem{zhang2019optimal}
Zhen Zhang, Mianzhi Wang, and Arye Nehorai,
\newblock ``Optimal transport in reproducing kernel {H}ilbert spaces: theory
  and applications,''
\newblock {\em IEEE Transactions on Pattern Analysis and Machine Intelligence},
  vol. 42, no. 7, pp. 1741–1754, 2020.

\bibitem{fukumizu2009kernel}
Kenji Fukumizu, Arthur Gretton, Gert~R. Lanckriet, Bernhard Sch{\"o}lkopf, and
  Bharath~K. Sriperumbudur,
\newblock ``Kernel choice and classifiability for {RKHS} embeddings of
  probability distributions,''
\newblock in {\em Advances in Neural Information Processing Systems}, 2009, pp.
  1750--1758.

\bibitem{sra2021LinearAlgebraApp}
Suvrit Sra,
\newblock ``Metrics induced by {Jensen-Shannon} and related divergences on
  positive definite matrices,''
\newblock {\em Linear Algebra and its Applications}, vol. 616, pp. 125--138,
  2021.

\bibitem{goodfellow2014gan}
Ian Goodfellow, Jean Pouget-Abadie, Mehdi Mirza, Bing Xu, David Warde-Farley,
  Sherjil Ozair, Aaron Courville, and Yoshua Bengio,
\newblock ``Generative adversarial nets,''
\newblock in {\em Advances in Neural Information Processing Systems},
  Z.~Ghahramani, M.~Welling, C.~Cortes, N.~Lawrence, and K.~Q. Weinberger, Eds.
  2014, vol.~27, Curran Associates, Inc.

\bibitem{endres2003ieeetransit}
D.~M. Endres and J.~E. Schindelin,
\newblock ``A new metric for probability distributions,''
\newblock {\em IEEE Transactions on Information Theory}, vol. 49, no. 7, pp.
  1858--1860, 2003.

\bibitem{rahimi2007random}
Ali Rahimi and Benjamin Recht,
\newblock ``Random features for large-scale kernel machines,''
\newblock in {\em Advances in Neural Information Processing Systems}, 2008, pp.
  1177--1184.

\bibitem{ermon2017infovae}
Shengjia Zhao, Jiaming Song, and Stefano Ermon,
\newblock ``{InfoVAE}: Information maximizing variational autoencoders,''
\newblock {\em ArXiv}, vol. abs/1706.02262, 2017.

\bibitem{principe2010itlbook}
Jose~C. Principe,
\newblock {\em Information Theoretic Learning: Renyi's Entropy and Kernel
  Perspectives},
\newblock Springer Publishing Company, Incorporated, 1st edition, 2010.

\bibitem{hoyos2021relevant}
J.~Hoyos-Osorio, A.~Alvarez-Meza, Genaro Daza-Santacoloma, A.~Orozco-Gutierrez,
  and Germ{\'a}n Castellanos-Dominguez,
\newblock ``Relevant information undersampling to support imbalanced data
  classification,''
\newblock {\em Neurocomputing}, vol. 436, pp. 136--146, 2021.

\bibitem{fernandez2008study}
Alberto Fern{\'a}ndez, Salvador Garc{\'\i}a, Mar{\'\i}a~Jos{\'e} del Jesus, and
  Francisco Herrera,
\newblock ``A study of the behaviour of linguistic fuzzy rule based
  classification systems in the framework of imbalanced data-sets,''
\newblock {\em Fuzzy Sets and Systems}, vol. 159, no. 18, pp. 2378--2398, 2008.

\bibitem{seiffert2010rusboost}
Chris Seiffert, Taghi~M. Khoshgoftaar, Jason Van~Hulse, and Amri Napolitano,
\newblock ``{RUSBoost}: A hybrid approach to alleviating class imbalance,''
\newblock {\em IEEE Transactions On Systems, Man, And CyberneticsPart A:
  Systems And Humans}, vol. 40, no. 1, pp. 185--197, 2010.

\bibitem{sun2020class}
Jie Sun, Hui Li, Hamido Fujita, Binbin Fu, and Wenguo Ai,
\newblock ``Class-imbalanced dynamic financial distress prediction based on
  {Adaboost-SVM} ensemble combined with {SMOTE} and time weighting,''
\newblock {\em Information Fusion}, vol. 54, pp. 128--144, 2020.

\bibitem{barandela2003new}
Ricardo Barandela, Rosa~Maria Valdovinos, and Jos{\'e}~Salvador S{\'a}nchez,
\newblock ``New applications of ensembles of classifiers,''
\newblock {\em Pattern Analysis \& Applications}, vol. 6, no. 3, pp. 245--256,
  2003.

\bibitem{raghuwanshi2019class}
Bhagat~Singh Raghuwanshi and Sanyam Shukla,
\newblock ``Class imbalance learning using underbagging based kernelized
  extreme learning machine,''
\newblock {\em Neurocomputing}, vol. 329, pp. 172--187, 2019.

\bibitem{wang2009diversity}
Shuo Wang and Xin Yao,
\newblock ``Diversity analysis on imbalanced data sets by using ensemble
  models,''
\newblock in {\em Computational Intelligence and Data Mining, 2009. CIDM'09.
  IEEE Symposium on}. IEEE, 2009, pp. 324--331.

\bibitem{sun2018imbalanced}
Jie Sun, Jie Lang, Hamido Fujita, and Hui Li,
\newblock ``Imbalanced enterprise credit evaluation with {DTE-SBD}: Decision
  tree ensemble based on {SMOTE} and bagging with differentiated sampling
  rates,''
\newblock {\em Information Sciences}, vol. 425, pp. 76--91, 2018.

\bibitem{CBUS3}
Wei-Chao Lin, Chih-Fong Tsai, Ya-Han Hu, and Jing-Shang Jhang,
\newblock ``Clustering-based undersampling in class-imbalanced data,''
\newblock {\em Information Sciences}, vol. 409, pp. 17--26, 2017.

\end{thebibliography}

\end{document}